\documentclass[runningheads]{llncs}

\usepackage{eccv}

\usepackage{eccvabbrv}

\usepackage{graphicx}
\usepackage{booktabs}

\usepackage[accsupp]{axessibility}  

\usepackage{hyperref}

\usepackage{orcidlink}

\usepackage{bm}
\usepackage{graphicx}
\usepackage{float}
\usepackage{wrapfig}
\usepackage{enumitem} 
\usepackage{algorithm}
\usepackage{setspace}
\usepackage{multicol}
\usepackage{multirow}
\usepackage{colortbl}
\usepackage{makecell}
\usepackage{pifont} 
\usepackage{amssymb}

\definecolor{cred}{HTML}{FF6B6B}
\definecolor{cyellow}{HTML}{FEC260}
\definecolor{cgreen}{HTML}{70AD47}
\definecolor{cblue}{HTML}{4D96FF}
\definecolor{cpurple}{HTML}{2A0944}
\definecolor{ggray}{RGB}{127,127,127}
\definecolor{aliceblue}{rgb}{0.94, 0.97, 1.0}

\makeatletter
\newcommand{\ssymbol}[1]{$^{\@fnsymbol{#1}}$}
\makeatother

\newcommand{\myparagraph}[1]{\textbf{#1}\hspace{1.8ex}}
\newcommand{\mysubparagraph}[1]{\textit{#1}\hspace{1.8ex}}

\newcommand{\pub}[1]{\color{gray}{\tiny{#1}}}

\begin{document}

\title{Local Action-Guided Motion Diffusion Model for Text-to-Motion Generation} 

\titlerunning{Local Action-Guided Motion Diffusion Model for Text-to-Motion Generation}

\author{Peng Jin\inst{1,2,3}\orcidlink{0000-0001-9287-6410} \and
Hao Li\inst{1,2,3}\orcidlink{0000-0002-3200-0270} \and
Zesen Cheng\inst{1,3} \and
Kehan Li\inst{1,3} \and
Runyi Yu\inst{1,3} \and
Chang Liu\inst{4\footnotemark[1]}\orcidlink{0000-0001-6747-0646} \and
Xiangyang Ji\inst{4}\orcidlink{0000-0002-7333-9975} \and
Li Yuan\inst{1,2,3\footnotemark[1]}\orcidlink{0000-0002-2120-5588} \and
Jie Chen\inst{1,2,3}\orcidlink{0000-0002-9765-4523} }

\authorrunning{Jin, Peng, et al.}

\institute{School of Electronic and Computer Engineering, Peking University, Shenzhen, China \and
Peng Cheng Laboratory, Shenzhen, China \and
AI for Science (AI4S)-Preferred Program, Peking University Shenzhen Graduate School, Shenzhen, China \and 
Department of Automation and BNRist, Tsinghua University, Beijing, China \\
\email{jp21@stu.pku.edu.cn, liuchang2022@tsinghua.edu.cn, yuanli-ece@pku.edu.cn}}

\maketitle

\renewcommand{\thefootnote}{\fnsymbol{footnote}}

\footnotetext[1]{Corresponding author: Li Yuan, Chang Liu.}

\begin{abstract}
  Text-to-motion generation requires not only grounding local actions in language but also seamlessly blending these individual actions to synthesize diverse and realistic global motions. However, existing motion generation methods primarily focus on the direct synthesis of global motions while neglecting the importance of generating and controlling local actions. In this paper, we propose the local action-guided motion diffusion model, which facilitates global motion generation by utilizing local actions as fine-grained control signals. Specifically, we provide an automated method for reference local action sampling and leverage graph attention networks to assess the guiding weight of each local action in the overall motion synthesis. During the diffusion process for synthesizing global motion, we calculate the local-action gradient to provide conditional guidance. This local-to-global paradigm reduces the complexity associated with direct global motion generation and promotes motion diversity via sampling diverse actions as conditions. Extensive experiments on two human motion datasets, \ie, HumanML3D and KIT, demonstrate the effectiveness of our method. Furthermore, our method provides flexibility in seamlessly combining various local actions and continuous guiding weight adjustment, accommodating diverse user preferences, which may hold potential significance for the community. The project page is available at \href{https://jpthu17.github.io/GuidedMotion-project/}{https://jpthu17.github.io/GuidedMotion-project/}.
\end{abstract}

\section{Introduction}
Human motion generation~\cite{plappert2018learning,ahuja2019language2pose,bhattacharya2021text2gestures} is a critical task in computer animation~\cite{tevet2022human,badler1993simulating}, with the primary objective of creating realistic and dynamic motions for virtual human characters. This technology finds widespread applications in multiple industries, such as entertainment, gaming, film production, virtual reality, and robotics. Recent developments in this field have introduced text-driven human motion generation techniques, enabling the synthesis of diverse human motion sequences based on natural language descriptions. However, text-driven human motion generation poses a series of challenges, requiring the alignment of local actions with language and the seamless integration of these individual actions to synthesize diverse and realistic global motions.

Existing text-to-motion generation methods~\cite{petrovich2022temos,ahn2018text2action,zhang2023remodiffuse,bhattacharya2021text2gestures} primarily focus on directly synthesizing global motions based on language instructions. Although these methods have shown promising advancements and achieved high accuracy, they come with limitations regarding the type of control they support over the motion results. For example, precisely expressing intricate trajectories, postures, and long motion sequences involving multiple actions is challenging using text prompts alone. Typically, generating a motion that faithfully corresponds to our mental imagery requires numerous iterations of editing a prompt, reviewing the resulting motion, and then adjusting the prompt accordingly.

\begin{figure*}[tbp]
\centering
\includegraphics[width=1.0\textwidth]{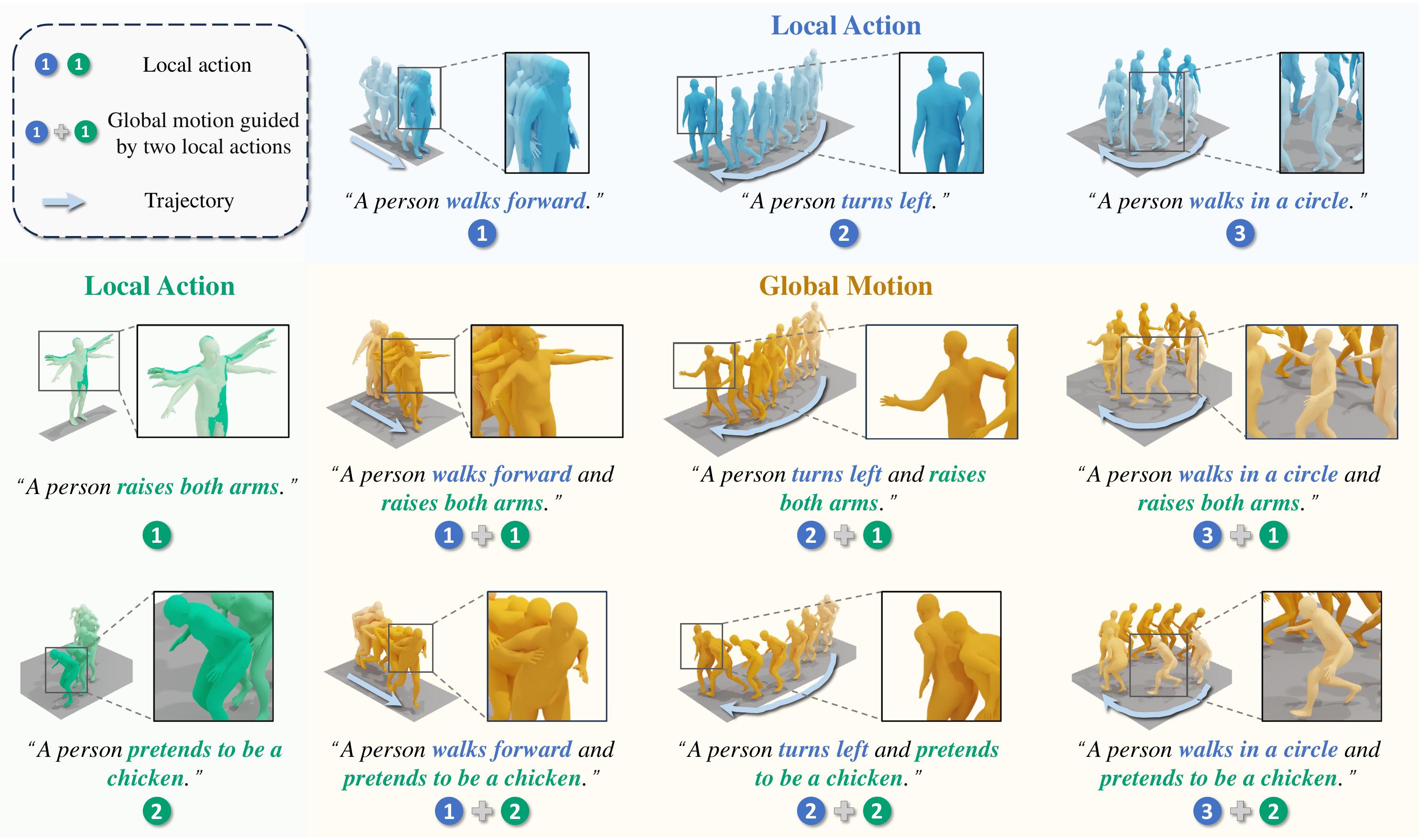}
\caption{\textbf{Generating motion with diverse local actions.} Different local actions correspond to distinct user preferences. Our method empowers users to combine preferred local actions freely, generating motions that align with their mental imagery. Furthermore, the combination of varied local actions enhances the motion diversity.}
\label{fig: fig1}
\end{figure*}

In this work, we propose to employ reference local actions as control signals in the global motion generation process. As illustrated in \cref{fig: fig1}, an overall motion comprises a sequence of local actions, such as ``\textit{walks forward}'' and ``\textit{raises both arms}''. These reference local actions can serve as control signals during the global motion generation process, facilitating the generation of global motions with similar characteristics, including movement trajectories and human body postures, to those of the local actions. More importantly, local action serves as a more intuitive control signal than text. Users can seamlessly combine their preferred local actions, exerting precise control over the resulting global motion to align with the characteristics of those chosen local actions.

To this end, we introduce GuidedMotion, a local action-guided motion diffusion model designed for controllable text-to-motion generation. Moreover, we provide an automatic local action sampling method, which deconstructs the original motion description into multiple local action descriptions and uses a text-to-motion model to generate the reference local actions. In practical applications, the same reference local action can be sampled multiple times to suit diverse user preferences, allowing users to conveniently select their preferred action from these choices. Subsequently, we leverage graph attention networks to estimate the guiding weight of each local motion in the overall motion synthesis. To enhance generation stability, we divide the motion diffusion process for synthesizing global motion into three stages: (i) In the initial diffusion stage, we de-noise the Gaussian noise based on the original motion description to provide a good initial value for the subsequent stage. (ii) In the second diffusion stage, we apply local-action gradients based on the energy function~\cite{zhao2022egsde} to offer conditional guidance for aligning the generated motion with the characteristics of the reference local actions. (iii) In the final diffusion stage, we fine-tune the generated results further to conform to the original motion description, rather than solely adhering to a reference local action.

The proposed GuidedMotion has three distinct advantages: \textbf{First}, compared to the direct generation of global motion, our local-to-global paradigm, leveraging local actions as a prior, simplifies the complexity associated with global motion generation, especially when generating complex motions with multiple local actions. \textbf{Second}, through the automatic sampling of diverse local actions, our method has the capability to generate a variety of motions to suit different user preferences. \textbf{Third}, our method provides flexibility in adjusting the guiding weight of each local action, enabling fine-grained and continuous control over global motion, \eg, the control of movement trajectories and human body postures. Extensive experiments on two datasets for text-to-motion generation, including HumanML3D~\cite{guo2022generating} and KIT~\cite{plappert2016kit}, demonstrate the advantages of GuidedMotion. The main contributions are summarized as follows:
\begin{itemize}
    \item We propose local action-guided motion synthesis for fine-grained controllable text-to-motion generation. It allows users to seamlessly combine their preferred local actions, enabling them to exert control over the resulting global motion to align with the characteristics of their chosen local actions.

    \item The proposed local-to-global paradigm, utilizing local actions as a prior, reduces the complexity associated with direct global motion generation. Experimental results demonstrate that our method has an advantage in generating complex motions comprising multiple local actions.

    \item More encouragingly, our method allows for continuous guiding weight adjustment, facilitating the refinement of the results to match the preferences of users, which may hold potential significance for the community.
\end{itemize}

\section{Related Work}
\noindent \myparagraph{Diffusion Models.}
Diffusion models~\cite{sohl2015deep,ho2020denoising,dhariwal2021diffusion,jeong2023zero,song2020denoising,zhang2023diffcollage,huang2023diffusion}, rooted in thermodynamics, utilize a stochastic diffusion process to complete the generation task. In recent years, diffusion models have exhibited potential across diverse tasks, including image generation~\cite{ho2020denoising,song2020denoising,dhariwal2021diffusion,ho2022classifier,wang2022zero}, natural language generation~\cite{austin2021structured,gong2022diffuseq}, and visual tasks~\cite{cheng2023parallel}. Some other works~\cite{jin2023diffusionret} have applied diffusion models to cross-modal retrieval~\cite{jin2023video}. Inspired by the success of diffusion generative models, some works~\cite{tevet2022human,karunratanakul2023guided,xu2023interdiff,yuan2023physdiff} have explored the application of diffusion models in human motion tasks~\cite{chen2023humanmac,barquero2023belfusion,lin2023motionx}. Although existing text-to-motion generation methods have shown promising advancements and achieved high accuracy, they come with limitations regarding the type of control they support over the motion results. In this paper, we propose to employ reference local actions as fine-grained control signals in the global motion generation process.

\noindent \myparagraph{Text-driven Human Motion Generation.}
The goal of text-driven human motion generation~\cite{zhu2023human,delmas2022posescript,he2022nemf} is to generate human motion based on text descriptions. Due to the user-friendly and convenient nature of natural language~\cite{jin2022expectation}, text-driven human motion generation has garnered significant attention. Recently, motion generation methods can be classified into three categories: joint-latent models~\cite{ahuja2019language2pose,petrovich2022temos}, such as TEMOS~\cite{petrovich2022temos}, which typically learn a motion variational autoencoder and a text variational autoencoder; the second category~\cite{chen2022executing,zhang2022motiondiffuse,shafir2023human,jin2023act}, such as MDM~\cite{tevet2022human}, introduces a conditional diffusion model for human motion generation; and the last category~\cite{zhang2023motiongpt,jiang2023motiongpt}, such as T2M-GPT~\cite{zhang2023t2m}, utilizes generative pre-trained transformer for motion generation. In this work, leveraging the iterative refinement of diffusion models, we employ the diffusion model method to enhance control over the motion generation process.

\noindent \myparagraph{Controllable Human Motion Generation.}
Controllable human motion generation~\cite{karunratanakul2023guided,xie2023omnicontrol,zhai2023language,yu2023point} aims to generate motions following designated control signals, offering enhanced interactivity and interpretability to humans. Existing methods predominantly focus on controlling trajectory and key points within the diffusion process through techniques such as imputation and inpainting. However, these low-level control signals lack the capability for high-level control over motions, such as adjusting the amplitude of arms. What is worse, existing methods lack support for continuous motion adjustment, limiting the ability to refine motions until they align with the expectations of users. In contrast, our method employs local actions with high-level semantics as control signals, enabling not only trajectory control but also manipulation of human body postures.

\begin{figure*}[tbp]
\centering
\includegraphics[width=1\textwidth]{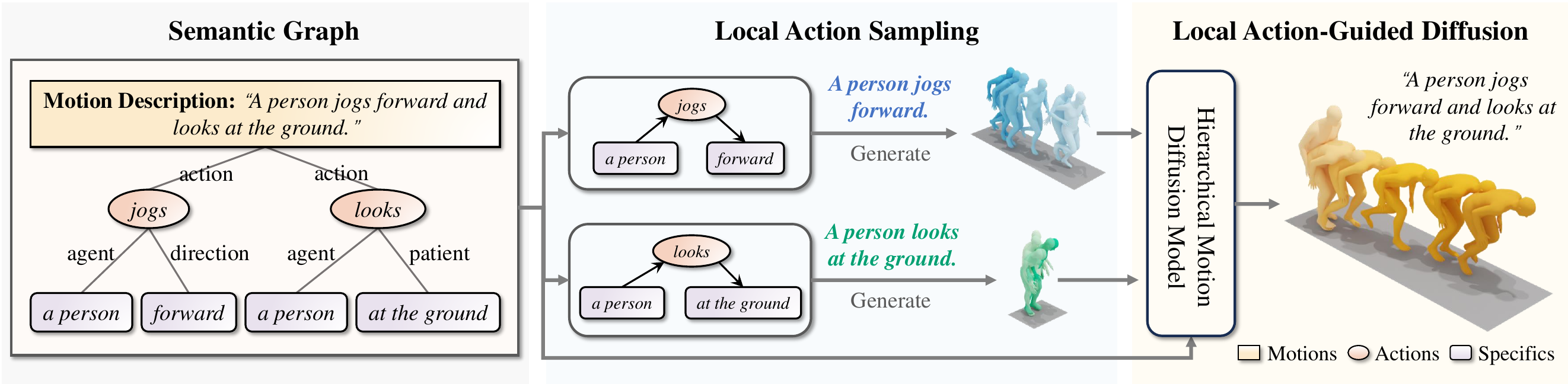}
\caption{\textbf{The overall framework of GuidedMotion for controllable text-to-motion generation.} We propose to employ reference local actions as control signals in the global motion generation process. To automatically obtain these local actions, we deconstruct the original motion description into multiple local action descriptions and utilize a text-to-motion model to generate these local actions.}
\label{fig: fig2}
\end{figure*}

\section{Methodology}
In this work, we tackle the challenges associated with controllable text-driven human motion generation. Concretely, given a motion description and other fine-grained control signals, such as reference local actions, our goal is to synthesize a human motion sequence $\bm{x}^{1:L}=\{x^i\}_{i=1}^{L}$ of length $L$.

\subsection{Automatic Local Action Sampling}
Local actions can be accessible from the repository, enabling users to choose their preferred local action as the control signal for generating the global motion. Moreover, we provide an automatic local action sampling method, which deconstructs the original motion description into multiple local action descriptions and utilizes a text-to-motion model to generate these local actions.

\noindent \myparagraph{Semantic Graph Parsing.} As shown in \cref{fig: fig2}, motion descriptions inherently exhibit hierarchical structures, represented as hierarchical graphs comprising three types of abstract nodes: motions, actions, and specifics. Concretely, the complete sentence describes the global motion, encompassing multiple actions, for example, ``\textit{jogs}'' and ``\textit{looks}'' in \cref{fig: fig2}. Each action includes various specifics, which serve as its attributes, such as the agent and patient of the action. 

To obtain actions, action attributes, and the semantic role of each attribute in relation to the corresponding action, we employ a semantic parser for motion descriptions based on a semantic role parsing toolkit~\cite{shi2019simple,jin2023act,chen2020fine}. In practice, we extract three types of nodes (motions, actions, and specifics) and twelve types of edges to represent various associations among the nodes. For further details about semantic graph parsing, please refer to our supplementary material.

\noindent \myparagraph{Local Action Sampling.} Given the semantic graph, we create a local action description for each local action by considering each action node and its associated specific nodes. Subsequently, We employ a text-to-motion generation model, \ie, MLD~\cite{chen2022executing}, to generate local actions based on these local action descriptions. To further enrich the variety of local actions, the local action descriptions can be expanded using large language models, such as GPT~\cite{brown2020language} and LLaMA~\cite{touvron2023llama,touvron2023llama2}.

\subsection{Local Action Diffusion Guidance}
Following previous works~\cite{chen2022executing,jin2023act}, we encode the motion sequence $\bm{x}$ into the latent space $\bm{z}$ utilizing the variational autoencoder~\cite{kingma2013auto}. Subsequently, we employ diffusion models to learn the noise component $\boldsymbol{\epsilon}$ at every noise level $t$.

\noindent \myparagraph{Energy Diffusion Guidance.} In accordance with score theory~\cite{song2019generative,song2020score,yu2023freedom}, the core objective of conditional diffusion models~\cite{ho2020denoising} is to estimate the score function $\nabla_{\bm{z}_t} \log{p(\bm{z}_t|\bm{c})}$, where $\bm{c}$ is the condition. The reverse process of conditional diffusion models is formulated as:
\begin{equation}
\bm{z}_{t-1}=(1+\frac{1}{2}{\beta}_t)\bm{z}_t + {\beta}_t \nabla_{\bm{z}_t} \log{p(\bm{z}_t|\bm{c})}+\sqrt{{\beta}_t}\boldsymbol{\epsilon},
\end{equation}
where $\boldsymbol{\epsilon}$ is a noise sampled from the standard Gaussian distribution $\mathcal{N}(\mathbf{0}, \mathbf{I})$. $\beta_t\in\mathbb{R}$ is a pre-defined step size which gradually increases. Based on Bayesian formula $p(\bm{z}_t|\bm{c})=\frac{p(\bm{c}|\bm{z}_t)p(\bm{z}_t)}{p(\bm{c})}$, we rewrite the conditional score function as:
\begin{equation}
    \nabla_{\bm{z}_t}\log p(\bm{z}_t|\bm{c}) = \nabla_{\bm{z}_t}\log p(\bm{z}_t) + \nabla_{\bm{z}_t}\log p(\bm{c}|\bm{z}_t),
\end{equation}
where the correction gradient $\nabla_{\bm{z}_t}\log p(\bm{c}|\bm{z}_t)$ holds paramount significance in the conditional diffusion models. However, the correction gradient $\nabla_{\bm{z}_t}\log p(\bm{c}|\bm{z}_t)$ is hard to measure directly. Following previous works~\cite{zhao2022egsde,yu2023freedom}, we employ the energy function~\cite{lecun2006tutorial} to formulate the correction term as:
\begin{equation}
    p(\bm{c}|\bm{z}_t)=\frac{\exp\left(-\mathcal{E}(\bm{c}, \bm{z}_t)\right)}{\int_{\bm{c}\in\mathcal{C}}\exp\left(-\mathcal{E}(\bm{c}, \bm{z}_t)\right)},
    \label{eq: energy}
\end{equation}
where $\mathcal{C}$ denotes the domain of the condition $\bm{c}$. With \cref{eq: energy}, the correction gradient can be estimated by the gradient of the energy function $\mathcal{E}(\bm{c}, \bm{z}_t)$, \ie, $\nabla_{\bm{z}_t}\log p(\bm{c}|\bm{z}_t)\propto -\nabla_{\bm{z}_t}\mathcal{E}(\bm{c}, \bm{z}_t)$. Therefore, the reverse process of conditional diffusion models can be rewritten as:
\begin{equation}
    \bm{z}_{t-1}=\tilde{\bm{z}}_{t-1} - \lambda_t\nabla_{\bm{z}_t}\mathcal{E}(\bm{c}, \bm{z}_t),
    \label{eq: energy1}
\end{equation}
where $\tilde{\bm{z}}_{t-1}=(1+\frac{1}{2}{\beta}_t)\bm{z}_t + {\beta}_t \nabla_{\bm{z}_t} \log{p(\bm{z}_t)}+\sqrt{{\beta}_t}\boldsymbol{\epsilon}$ is the original reverse process of unconditional diffusion models. In essence, $\lambda_t$ is the guiding weight, which represents the learning rate of the correction. When there are multiple conditions in the reverse process of diffusion models, \cref{eq: energy1} is reformulated as:
\begin{equation}
    \bm{z}_{t-1}=\tilde{\bm{z}}_{t-1} - \sum^K_{k=1} \lambda_t^k \nabla_{\bm{z}_t}\mathcal{E}(\bm{c}^k, \bm{z}_t),
\end{equation}
where $K$ represents the number of guidance terms.

In this work, the condition $\bm{c}$ is the motion latent embeddings of local actions. The number $K$ of guidance local actions is determined based on semantic parsing of the input motion description. To achieve the goal of the diffusion guidance, the energy function $\mathcal{E}(\bm{c}, \bm{z}_t)$ should meet all the following criteria: (i) if $\bm{z}_t$ is a better match with $\bm{c}$, then $\mathcal{E}(\bm{c}, \bm{z}_t)$ is smaller; (ii) If $\bm{z}_t$ perfectly conforms to the constraint set by $\bm{c}$, then $\mathcal{E}(\bm{c}, \bm{z}_t)$ should be zero. 

Note that anything satisfying the above conditions can be employed as the energy function $\mathcal{E}(\bm{c}, \bm{z}_t)$, such as the Gram matrix~\cite{johnson2016perceptual} distance and the embedding distance. For simplicity in implementation, we utilize the $\ell_2$ distance of latent embeddings as the energy function in practice.

\begin{figure*}[tbp]
\centering
\includegraphics[width=1\textwidth]{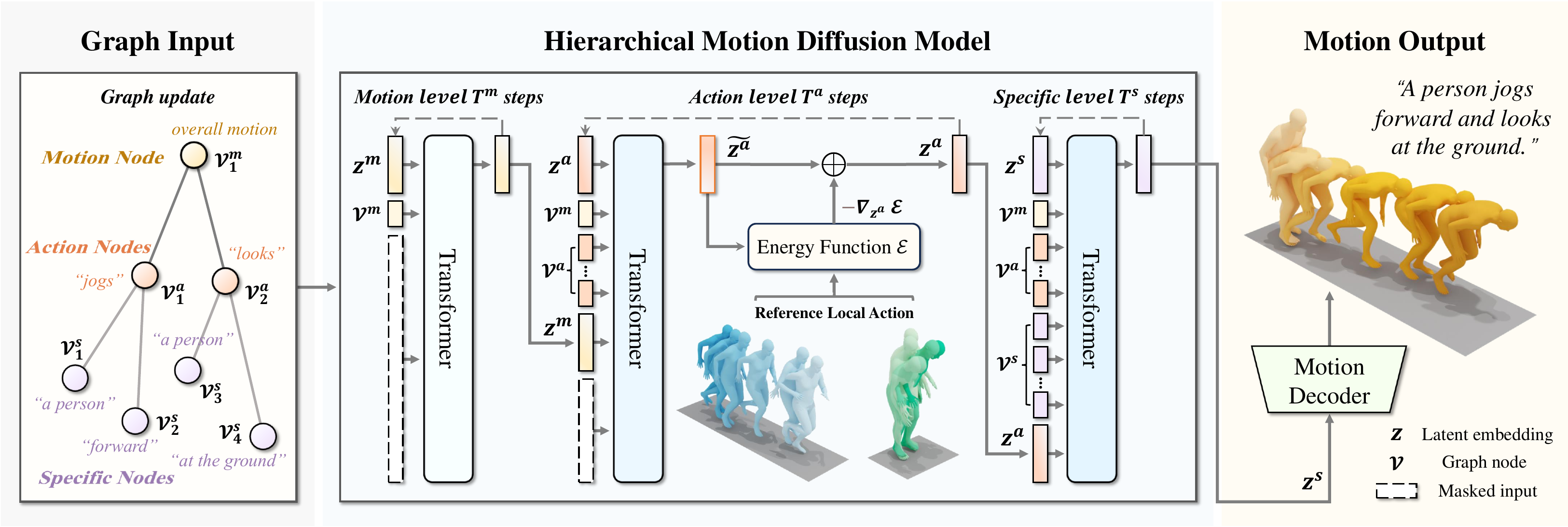}
\caption{\textbf{The model architecture of the hierarchical motion diffusion model.} Utilizing the semantic graph as input, the hierarchical diffusion model dissects the text-to-motion diffusion process into three semantic levels, which correspond to capturing the overall motion, local actions, and action specifics. To enhance generation stability, we exclusively implement local action guidance at the action level.}
\label{fig: fig3}
\end{figure*}

\noindent \myparagraph{Guiding Weight Estimation.} 
As illustrated in \cref{fig: fig3}, the interactions among different levels in the semantic graph elucidate the characteristics of local actions and how these local actions contribute to the overall motion. Drawing inspiration from this insight, we employ graph attention networks~\cite{velivckovic2017graph}~(GAT) to model the guiding weights in the local action-guided motion diffusion model.

We leverage the text encoder of CLIP~\cite{radford2021learning} to initialize the representation of graph nodes. To represent the global motion node $\bm{v}^{m}$, we utilize the [CLS] token to encapsulate the overall motion within the description. For the action node $\bm{v}^{a}$, we adopt the token corresponding to the verb as the representation. In the case of the specific node $\bm{v}^{s}$, we employ mean-pooling over tokens of each word in the attribute phrase to represent every action detail of the motion.

Given the initialized nodes $\bm{v}=\{\bm{v}^{m}, \bm{v}^{a}, \bm{v}^{s}\}$, we utilize a shared projection matrix $\bm{W} \in \mathbb{R}^{D\times D}$, where $D$ represents the dimension of node representation. This matrix transforms $\bm{v}$ into higher-level embeddings $\bm{h}=\{\bm{h}^{m}, \bm{h}^{a}, \bm{h}^{s}\}$. For each pair $\{\bm{h}_i, \bm{h}_j\}$ of connected nodes, we concatenate the node $\bm{h}_i\in \mathbb{R}^{D}$ with its neighbor node $\bm{h}_j\in \mathbb{R}^{D}$, creating the input data $\tilde{\bm{h}}_{ij}=[\bm{h}_i, \bm{h}_j]\in \mathbb{R}^{2D}$ of the graph attention module, which is formulated as:
\begin{equation}
\tilde{\bm{h}}_{ij}=[\bm{h}_i, \bm{h}_j]=[\bm{W}\bm{v}_i, \bm{W}\bm{v}_j].
\end{equation}

The semantic graph comprises multiple types of edges. To avoid over-fitting to infrequent edge types, we employ a shared transformation matrix $\bm{M} \in \mathbb{R}^{2D\times 1}$ that applies to all edge types, and a relationship embedding matrix $\bm{M_{r}} \in \mathbb{R}^{2D\times N}$ that is specific for different edges to represent multi-relational weights, where $N$ denotes the number of edge types. The attention coefficient $\tilde{e}_{ij}$ is formulated as:
\begin{equation}
\begin{aligned} 
e_{ij} = \sigma(\bm{M}^\top&\tilde{\bm{h}}_{ij}) + \sigma(\bm{R}_{ij}\bm{M_{r}}^\top\tilde{\bm{h}}_{ij}),\\
\tilde e_{ij} =& \frac{\textrm{exp}(e_{ij})}{\sum_{k\in \mathbb{N}_i}\textrm{exp}(e_{ik})},
\end{aligned} 
\end{equation}
where $\sigma$ refers to a nonlinear function, \ie LeakyReLU~\cite{maas2013rectifier} with a negative input slope set to 0.2. $\bm{R}_{ij} \in \mathbb{R}^{1\times N}$ denotes a one-hot vector denoting the type of edge between node $i$ and $j$. $\mathbb{N}_i$ is the set of neighborhood nodes of node $i$. Finally, we use the attention coefficient $\tilde e$ as the guiding weight $\lambda$, which is formulated as:
\begin{equation}
\lambda_t^k = \rho_t \tilde{e}^k,\ \ \tilde{e}^k \in \big\{(\bm{u},\bm{v}^m) \big| \bm{u} \in \{ \bm{v}^a_k\}_{k=1}^K \big\},
\label{eq: energy2}
\end{equation}
where $\rho_t$ is a predefined parameter used to amplify or reduce the guiding strength. $\tilde{e}^k$ is the attention coefficient corresponding to the $k_{th}$ reference local action.

\subsection{Hierarchical Motion Diffusion Model}
To enhance generation stability, we decompose the diffusion process into three semantic levels and build three transformer-based denoising networks, which correspond to motions, actions, and specifics. The motion level provides a good initial value for the subsequent semantic levels. Subsequently, we exclusively implement local action guidance at the action level. Finally, at the specific level, we further refine the generated results to match the original motion description, rather than solely adhering to a reference local action.

\noindent \myparagraph{Motion Variational Autoencoder.}
Following previous works~\cite{petrovich2022temos,zhang2023t2m}, we encode the motion into the latent space using a motion variational autoencoder~\cite{kingma2013auto}~(VAE). Specifically, we construct the motion encoder and decoder based on the transformer~\cite{vaswani2017attention,petrovich2021action}. For the motion encoder, we utilize $Q$ learnable query tokens along with the motion sequence $\bm{x}^{1:L}=\{x^i\}_{i=1}^{L}$ as inputs to generate motion latent embeddings $\bm{z}\in \mathbb{R}^{Q\times D'}$, where $D'$ is the dimension of latent representation. For the motion decoder, we input the latent embeddings $\bm{z}\in \mathbb{R}^{Q\times D'}$ and the motion query tokens to generate a human motion sequence. 

Corresponding to the three-level structure of the hierarchical motion diffusion model, we encode human motion sequences independently into three latent representation spaces: $\bm{z}^m\in \mathbb{R}^{Q^m\times D'}$, $\bm{z}^a\in \mathbb{R}^{Q^a\times D'}$, and $\bm{z}^s\in \mathbb{R}^{Q^s\times D'}$. To generate motion progressively from coarse to fine, we gradually increase the number of learnable query tokens, \ie, $Q^m\leq Q^a \leq Q^s$.

\noindent \myparagraph{Hierarchical Motion Diffusion.} We utilize the semantic graph as the input for the hierarchical diffusion model. The node embeddings ${\mathcal{V}}$ are formulated as:
\begin{equation}
{\mathcal{V}}_{i}=\sigma'\big(\sum_{j\in \mathbb{N}_i} \tilde e_{ij} \bm{h}_{j}\big) + \bm{v}_{i},
\end{equation}
where $\sigma'$ is a nonlinear function. Following graph attention networks~\cite{velivckovic2017graph}~(GAT), we adopt ELU~\cite{clevert2015fast} as the nonlinear function $\sigma'$ and apply skip connection~\cite{ronneberger2015u,he2016deep} to mitigate over-smoothing~\cite{yang2020revisiting} in graph networks.

\begin{figure*}[tbp]
\centering
\includegraphics[width=1.\textwidth]{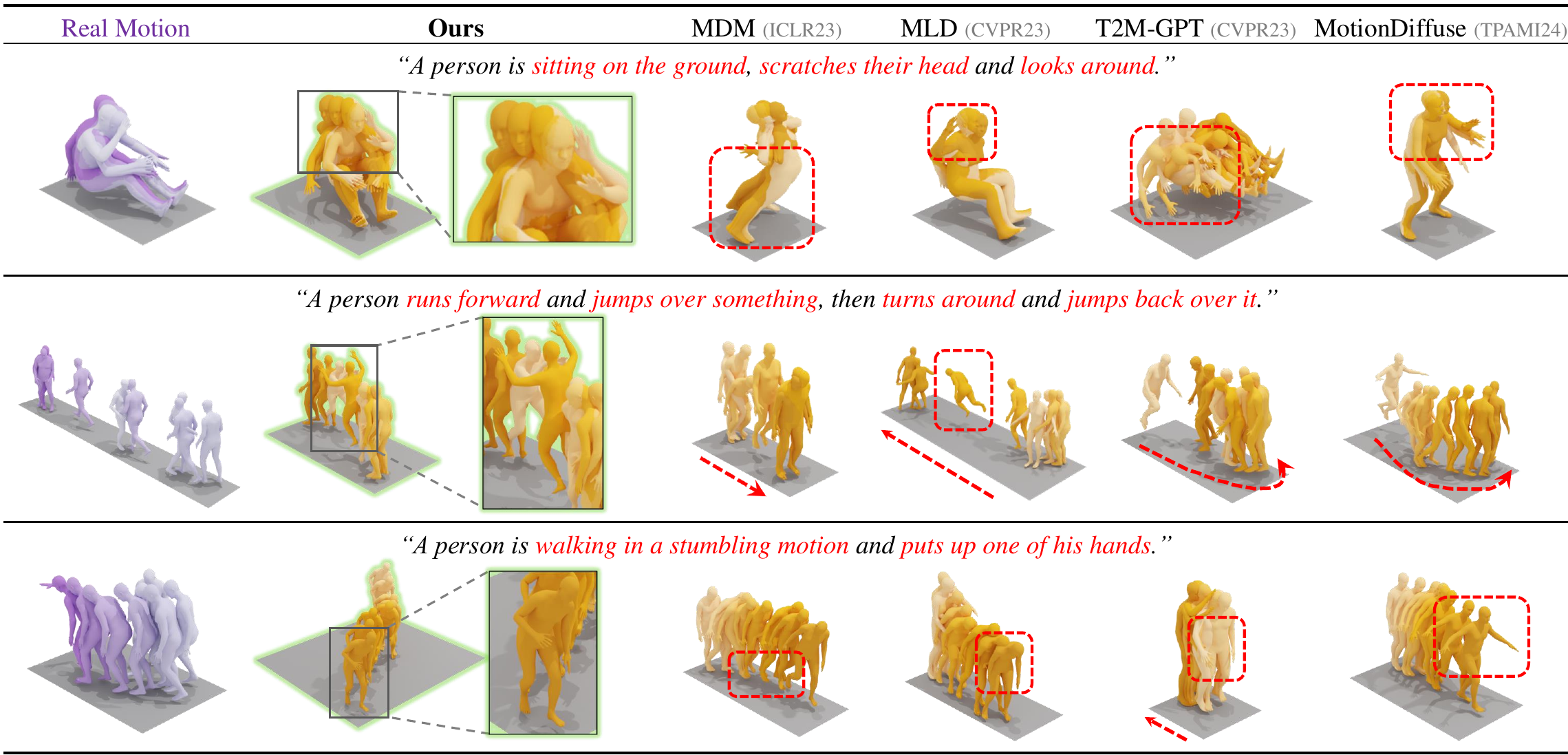}
\caption{\textbf{Qualitative comparisons.} The darker colors indicate the later in time. The motions generated by our method closely align with the descriptions, outperforming others that exhibit degraded motions or improper semantics.}
\label{fig:qualitative comparison}
\end{figure*}

\begin{table*}[t]
\centering
\caption{\textbf{Comparisons to current state-of-the-art methods on the HumanML3D test set.} ``$\uparrow$'' denotes that higher is better. ``$\downarrow$'' denotes that lower is better. ``$\rightarrow$'' denotes that results are better if the metric is closer to the real motion. We repeat all the evaluations 20 times and report the average with a 95\% confidence interval. \textbf{Bold} and \underline{underlined} indicate the best and second-best results, respectively.}
\label{tab:humanml3d}
\resizebox{1.\linewidth}{!}
{
\begin{tabular}{lccccccc}
\toprule[1.pt]
\multirow{2}{*}{{Methods}} &\multicolumn{3}{c}{{R-Precision $\uparrow$}} & \multirow{2}{*}{{FID $\downarrow$}} & \multirow{2}{*}{{MM-Dist $\downarrow$}} & \multirow{2}{*}{{Diversity $\rightarrow$}} & \multirow{2}{*}{{MModality $\uparrow$}}\\
\cmidrule(rl){2-4}
  & Top-1 & Top-2 & Top-3 \\ \midrule
Real Motion & $0.511^{\pm.003}$ & $0.703^{\pm.003}$ & $0.797^{\pm.002}$ & $0.002^{\pm.000}$ & $2.974^{\pm.008}$ & $9.503^{\pm.065}$ & - \\ \midrule
Hier~\cite{ghosh2021synthesis}~\pub{ICCV21} & $0.301^{\pm.002}$ & $0.425^{\pm.002}$ & $0.552^{\pm.004}$ & $6.532^{\pm.024}$ & $5.012^{\pm.018}$ & $8.332^{\pm.042}$ & - \\
TEMOS~\cite{petrovich2022temos}~\pub{ECCV22} & $0.424^{\pm.002}$ & $0.612^{\pm.002}$ & $0.722^{\pm.002}$ & $3.734^{\pm.028}$ & $3.703^{\pm.008}$ & $8.973^{\pm.071}$ & $0.368^{\pm.018}$ \\
TM2T~\cite{guo2022tm2t}~\pub{ECCV22} & $0.424^{\pm.003}$ & $0.618^{\pm.003}$ & $0.729^{\pm.002}$ & $1.501^{\pm.017}$ & $3.467^{\pm.011}$ & $8.589^{\pm.076}$ & $2.424^{\pm.093}$ \\
T2M~\cite{guo2022generating}~\pub{CVPR22} & $0.457^{\pm.002}$ & $0.639^{\pm.003}$ & $0.740^{\pm.003}$ & $1.067^{\pm.002}$ & $3.340^{\pm.008}$ & $9.188^{\pm.002}$ & $2.090^{\pm.083}$ \\
MotionDiffuse~\cite{zhang2022motiondiffuse}~\pub{TPAMI24} & $0.491^{\pm.001}$ & ${0.681^{\pm.001}}$ & ${0.782^{\pm.001}}$ & $0.630^{\pm.001}$ & ${3.113^{\pm.001}}$ & ${9.410^{\pm.049}}$ & $1.553^{\pm.042}$ \\
MDM~\cite{tevet2022human}~\pub{ICLR23} & $0.320^{\pm.005}$ & $0.498^{\pm.004}$ & $0.611^{\pm.007}$ & $0.544^{\pm.044}$ & $5.566^{\pm.027}$ & $\underline{9.559^{\pm.086}}$ & $\bm{2.799^{\pm.072}}$ \\
MLD~\cite{chen2022executing}~\pub{CVPR23} & $0.481^{\pm.003}$ & $0.673^{\pm.003}$ & $0.772^{\pm.002}$ & $0.473^{\pm.013}$ & $3.196^{\pm.010}$ & $9.724^{\pm.082}$ & $2.413^{\pm.079}$ \\
Fg-T2M~\cite{wang2023fg}~\pub{ICCV23} & $0.492^{\pm.002}$ & $0.683^{\pm.003}$ & $0.783^{\pm.002}$ & $0.243^{\pm.019}$ & ${3.109^{\pm.007}}$ & $9.278^{\pm.072}$ & $1.614^{\pm.049}$ \\
MotionGPT~\cite{jiang2023motiongpt}~\pub{NeurIPS23} & $0.492^{\pm.003}$ & $0.681^{\pm.003}$ & $0.778^{\pm.002}$ & $0.232^{\pm.008}$ & $3.096^{\pm.008}$ & $\bm{9.528^{\pm.071}}$ & $2.008^{\pm.084}$ \\
T2M-GPT~\cite{zhang2023t2m}~\pub{CVPR23} & ${0.491^{\pm.003}}$ & $0.680^{\pm.003}$ & $0.775^{\pm.002}$ & ${0.116^{\pm.004}}$ & $3.118^{\pm.011}$ & ${9.761^{\pm.081}}$ & $1.856^{\pm.011}$ \\ 
GraphMotion~\cite{jin2023act}~\pub{NeurIPS23} & $\underline{0.504^{\pm.003}}$ & $\bm{0.699^{\pm.002}}$ & ${0.785^{\pm.002}}$ & ${0.116^{\pm.007}}$ & ${3.070^{\pm.008}}$ & ${9.692^{\pm.067}}$ & $\underline{2.766^{\pm.096}}$\\
ReMoDiffuse~\cite{zhang2023remodiffuse}~\pub{ICCV23} & $\bm{0.510^{\pm.005}}$ & $\underline{0.698^{\pm.006}}$ & $\bm{0.795^{\pm.004}}$ & $\underline{0.103^{\pm.004}}$ & $\bm{2.974^{\pm.016}}$ & ${9.018^{\pm.075}}$ & $1.795^{\pm.043}$\\
\midrule
\rowcolor{aliceblue!60} GuidedMotion (Ours) & ${0.503^{\pm.002}}$ & ${0.691^{\pm.002}}$ & $\underline{0.788^{\pm.002}}$ & $\bm{0.057^{\pm.006}}$ & $\underline{3.040^{\pm.012}}$ & ${9.864^{\pm.077}}$ & ${2.473^{\pm.096}}$\\
\bottomrule[1.pt]
\end{tabular}
}
\end{table*}

\begin{table*}[t]
\centering
\caption{\textbf{Comparisons to other methods on the KIT test set.} We repeat all the evaluations 20 times and report the average with a 95\% confidence interval. \textbf{Bold} and \underline{underlined} indicate the best and second-best results, respectively.}
\label{tab:kit}
\resizebox{1.\linewidth}{!}{
\begin{tabular}{lccccccc}
\toprule[1.pt]
\multirow{2}{*}{{Methods}} &\multicolumn{3}{c}{{R-Precision $\uparrow$}} & \multirow{2}{*}{{FID $\downarrow$}} & \multirow{2}{*}{{MM-Dist $\downarrow$}} & \multirow{2}{*}{{Diversity $\rightarrow$}} & \multirow{2}{*}{{MModality $\uparrow$}}\\
\cmidrule(rl){2-4}
  & Top-1 & Top-2 & Top-3 \\ \midrule
Real Motion & $0.424^{\pm.005}$ & $0.649^{\pm.006}$ & $0.779^{\pm.006}$ & $0.031^{\pm.004}$ & $2.788^{\pm.012}$ & $11.08^{\pm.097}$ & - \\ \midrule
Hier~\cite{ghosh2021synthesis}~\pub{ICCV21} & $0.255^{\pm.006}$ & $0.432^{\pm.007}$ & $0.531^{\pm.007}$ & $5.203^{\pm.107}$ & $4.986^{\pm.027}$ & $9.563^{\pm.072}$ & $2.090^{\pm.083}$ \\
TEMOS~\cite{petrovich2022temos}~\pub{ECCV22} & $0.353^{\pm.006}$ & $0.561^{\pm.007}$ & $0.687^{\pm.005}$ & $3.717^{\pm.051}$ & $3.417^{\pm.019}$ & $10.84^{\pm.100}$ & $0.532^{\pm.034}$ \\
TM2T~\cite{guo2022tm2t}~\pub{ECCV22} & $0.280^{\pm.005}$ & $0.463^{\pm.006}$ & $0.587^{\pm.005}$ & $3.599^{\pm.153}$ & $4.591^{\pm.026}$ & $9.473^{\pm.117}$ & ${3.292^{\pm.081}}$ \\
T2M~\cite{guo2022generating}~\pub{CVPR22} & $0.370^{\pm.005}$ & $0.569^{\pm.007}$ & $0.693^{\pm.007}$ & $2.770^{\pm.109}$ & $3.401^{\pm.008}$ & $10.91^{\pm.119}$ & $1.482^{\pm.065}$ \\
MotionDiffuse~\cite{zhang2022motiondiffuse}~\pub{TPAMI24} & ${0.417^{\pm.004}}$ & $0.621^{\pm.004}$ & $0.739^{\pm.004}$ & $1.954^{\pm.062}$ & $\underline{2.958^{\pm.005}}$ & $\bm{11.10^{\pm.143}}$ & $0.730^{\pm.013}$ \\
Fg-T2M~\cite{wang2023fg}~\pub{ICCV23} & ${0.418^{\pm.005}}$ & $0.626^{\pm.004}$ & $0.745^{\pm.004}$ & $0.571^{\pm.047}$ & ${3.114^{\pm.015}}$ & ${10.93^{\pm.083}}$ & $1.019^{\pm.029}$ \\
T2M-GPT~\cite{zhang2023t2m}~\pub{CVPR23} & $0.416^{\pm.006}$ & $0.627^{\pm.006}$ & $0.745^{\pm.006}$ & $0.514^{\pm.029}$ & ${3.007^{\pm.023}}$ & $10.92^{\pm.108}$ & $1.570^{\pm.039}$ \\ 
MDM~\cite{tevet2022human}~\pub{ICLR23} & $0.164^{\pm.004}$ & $0.291^{\pm.004}$ & $0.396^{\pm.004}$ & $0.497^{\pm.021}$ & $9.191^{\pm.022}$ & $10.85^{\pm.109}$ & $1.907^{\pm.214}$ \\
MLD~\cite{chen2022executing}~\pub{CVPR23} & $0.390^{\pm.008}$ & $0.609^{\pm.008}$ & $0.734^{\pm.007}$ & $0.404^{\pm.027}$ & $3.204^{\pm.027}$ & $10.80^{\pm.117}$ & $2.192^{\pm.071}$ \\
GraphMotion~\cite{jin2023act}~\pub{NeurIPS23} & $\underline{0.429^{\pm.007}}$ & $\underline{0.648^{\pm.006}}$ & $\bm{0.769^{\pm.006}}$ & ${0.313^{\pm.013}}$ & ${3.076^{\pm.022}}$ & $\underline{11.12^{\pm.135}}$ & $\underline{3.627^{\pm.113}}$\\
ReMoDiffuse~\cite{zhang2023remodiffuse}~\pub{ICCV23} & ${0.427^{\pm.014}}$ & ${0.641^{\pm.004}}$ & ${0.765^{\pm.055}}$ & $\bm{0.155^{\pm.006}}$ & $\bm{2.814^{\pm.012}}$ & ${10.80^{\pm.105}}$ & $1.239^{\pm.028}$\\
\midrule
\rowcolor{aliceblue!60} GuidedMotion (Ours) & $\bm{0.430^{\pm.006}}$ & $\bm{0.652^{\pm.005}}$ & $\underline{0.768^{\pm.005}}$ & $\underline{0.213^{\pm.017}}$ & ${3.034^{\pm.021}}$ & ${10.99^{\pm.101}}$ & $\bm{4.138^{\pm.145}}$\\
\bottomrule[1.pt]
\end{tabular}
}
\end{table*}

To improve generation stability, we partition the diffusion process into three semantic levels, aligning with motions, actions, and specifics. In the motion level model $\phi_{m}$, We employ the global motion node ${\mathcal{V}}^{m}$ as input to predict the noise component $\boldsymbol{\epsilon}^m$. The training objective for the motion level is formulated as:
\begin{equation}
\begin{aligned}
\mathcal{L}_{M} = \mathbb{E}_{\bm{z},\boldsymbol{\epsilon},t} \Big [\Vert \boldsymbol{\epsilon}^m - \phi_{m}(\bm{z}^m, t^m, {\mathcal{V}}^{m}) \Vert^2_2 \Big ].
\end{aligned}
\end{equation}

In the action level model $\phi_{a}$, we concatenate the action node ${\mathcal{V}}^{a}$, the motion node ${\mathcal{V}}^{m}$, and the generated result ${\bm{z}^m}$ from the motion level as the input. The training objective for the action level is formulated as:
\begin{equation}
\begin{aligned}
\mathcal{L}_{A} = \mathbb{E}_{\bm{z},\boldsymbol{\epsilon},t} \Big [\Vert \boldsymbol{\epsilon}^a - \phi_{a}(\bm{z}^a, t^a, [{\mathcal{V}}^{m},{\mathcal{V}}^{a},\bm{z}^{m}]) \Vert^2_2 \Big ].
\end{aligned}
\end{equation}

In the specific level model $\phi_{s}$, we utilize the results generated by the action level and nodes across all semantic levels to predict the noise component. The training objective for the specific level is formulated as:
\begin{equation}
\begin{aligned}
\mathcal{L}_{S} = \mathbb{E}_{\bm{z},\boldsymbol{\epsilon},t} \Big [\Vert \boldsymbol{\epsilon}^s - \phi_{s}(\bm{z}^s, t^s, [\mathcal{V}^{m},\mathcal{V}^{a},\mathcal{V}^{s},\bm{z}^{a}]) \Vert^2_2 \Big ].
\end{aligned}
\end{equation}

Finally, the total training objective is denoted as $\mathcal{L}=\mathcal{L}_{M}+\mathcal{L}_{A}+\mathcal{L}_{S}$. The output at the specific level is considered the final result, and the motion decoder is employed to decode the latent representation into the motion sequence.

\section{Experiments}
\noindent \myparagraph{Experimental Settings.}
\mysubparagraph{Datasets.}
We compare the proposed method with other methods on two commonly used public benchmarks: HumanML3D~\cite{guo2022generating} and KIT~\cite{plappert2016kit}. \textbf{HumanML3D}~\cite{guo2022generating} originates from and textually reannotates the HumanAct12~\cite{guo2020action2motion} and AMASS~\cite{mahmood2019amass} datasets. HumanML3D comprises 14,616 human motions and 44,970 text descriptions. \textbf{KIT}~\cite{plappert2016kit} contains 3,911 human motion sequences and 6,278 textual annotations.

\begin{table*}[t]
\centering
\caption{\textbf{Comparisons to other methods on the complex motion subset.} We filter the HumanML3D test set containing at least 3 local actions and 150 frames or more in length as a new test set to verify the ability to generate complex motions.}
\label{tab:complex}
\resizebox{1.\linewidth}{!}
{
\begin{tabular}{lccccccc}
\toprule[1.pt]
\multirow{2}{*}{{Methods}} &\multicolumn{3}{c}{{R-Precision $\uparrow$}} & \multirow{2}{*}{{FID $\downarrow$}} & \multirow{2}{*}{{MM-Dist $\downarrow$}} & \multirow{2}{*}{{Diversity $\rightarrow$}} & \multirow{2}{*}{{MModality $\uparrow$}}\\
\cmidrule(rl){2-4}
  & Top-1 & Top-2 & Top-3 \\ \midrule
Real Motion & $0.456^{\pm.002}$ & $0.640^{\pm.002}$ & $0.740^{\pm.002}$ & $0.002^{\pm.000}$ & $3.245^{\pm.009}$ & $8.738^{\pm.064}$ & - \\ \midrule
MDM~\cite{tevet2022human}~\pub{ICLR23} & $0.300^{\pm.008}$ & $0.473^{\pm.006}$ & $0.581^{\pm.011}$ & $0.579^{\pm.057}$ & $5.437^{\pm.041}$ & $\bm{8.987^{\pm.101}}$ & $\bm{2.808^{\pm.063}}$ \\
MLD~\cite{chen2022executing}~\pub{CVPR23} & $0.417^{\pm.006}$ & $0.603^{\pm.005}$ & $0.710^{\pm.006}$ & $0.783^{\pm.069}$ & $\bm{3.243^{\pm.013}}$ & $\underline{9.235^{\pm.164}}$ & $\underline{2.642^{\pm.118}}$ \\
T2M-GPT~\cite{zhang2023t2m}~\pub{CVPR23} & $\underline{0.431^{\pm.003}}$ & $\underline{0.612^{\pm.003}}$ & $\underline{0.712^{\pm.002}}$ & $\underline{0.314^{\pm.004}}$ & $3.448^{\pm.011}$ & ${9.277^{\pm.081}}$ & $2.125^{\pm.011}$ \\ 
\midrule
\rowcolor{aliceblue!60} GuidedMotion (Ours) & $\bm{0.451^{\pm.003}}$ & $\bm{0.635^{\pm.003}}$ & $\bm{0.732^{\pm.002}}$ & $\bm{0.144^{\pm.008}}$ & $\underline{3.447^{\pm.011}}$ & ${9.284^{\pm.057}}$ & $2.503^{\pm.113}$\\
\bottomrule[1.pt]
\end{tabular}
}
\end{table*}

\noindent \mysubparagraph{Metrics.}
Following previous works, we use the following five metrics to measure the performance of the model. (1) \textbf{R-Precision.} In the feature space of the pre-trained network introduced by T2M~\cite{guo2022generating}, motion-retrieval precision is determined by the matching accuracy of the top 1/2/3 text descriptions with a motion sequence and 32 text descriptions. (2) \textbf{Frechet Inception Distance~(FID).} We measure the distribution distance between generated and real motion using FID~\cite{heusel2017gans} on the extracted motion features~\cite{guo2022generating}. (3) \textbf{Multimodal Distance~(MM-Dist).} We calculate the average Euclidean distances between each text feature and the corresponding generated motion feature. (4) \textbf{Diversity.} All generated motions are randomly sampled into two equal-sized subsets. Motion features~\cite{guo2022generating} are then extracted, and the average Euclidean distances between the two subsets represent diversity. (5) \textbf{Multimodality (MModality).} For each text description, we generate 20 motion sequences, creating 10 pairs of motions. The average Euclidean distance between motion features is calculated for each pair. The result is the average across all text descriptions.

\noindent \mysubparagraph{Implementation details.}
For text representation, we employ a frozen text encoder from the CLIP-ViT-L-14~\cite{radford2021learning} model. The dimension of node representation $D$ is set to 768. The dimension of latent embedding $D'$ is set to 256. We set the token sizes $Q^m$ to 2, $Q^a$ to 4, and $Q^s$ to 8. The predefined parameter $\rho$ in \cref{eq: energy2} is set to 0.01. All our models are trained using the AdamW~\cite{kingma2014adam,loshchilov2017fixing} optimizer with a fixed learning rate of 1e-4. Training is performed on 4 Tesla V100 GPUs, with 128 samples on each GPU, resulting in a total batch size of 512. We keep running a similar number of iterations on different datasets. For the HumanML3D dataset, the model is trained for 6,000 epochs during the motion variational autoencoder stage and 3,000 epochs during the diffusion stage. In the case of the KIT dataset, the model is trained for 30,000 epochs during the motion variational autoencoder stage and 15,000 epochs during the diffusion stage.

\noindent \myparagraph{Comparisons to State-of-the-Art.}
We provide qualitative motion results in \cref{fig:qualitative comparison}. Compared to other methods, our method generates motions that match the text descriptions better and are more realistic. Moreover, we compare the proposed GuidedMotion with other methods on two benchmarks. \cref{tab:humanml3d} shows the results on the HumanML3D test set. \cref{tab:kit} presents the results on the KIT test set. Across both benchmarks, the proposed GuidedMotion, which allows for continuous refinement of motion results, achieves performance comparable to existing state-of-the-art methods that lack fine-grained control.

\begin{table}[t]
\begin{minipage}[t]{0.48\textwidth}
{
\centering
\caption{\textbf{Ablation study of each part on the HumanML3D test set.}}
\label{tab:ab1}
\resizebox{1.\linewidth}{!}{
\begin{tabular}{cccc|cc}
\toprule[1.pt]
{Motion} & {Action} & {Specific} & {Local Action} & {R-Precision} & \multirow{2}{*}{{FID $\downarrow$}} \\ 
{Level} & {Level} & {Level} & {Guidance} & Top-3 $\uparrow$  \\ \midrule
 {\checkmark} & & & & $0.760^{\pm.003}$ & $0.186^{\pm.011}$  \\
 \checkmark & {\checkmark} & & & $0.771^{\pm.003}$ & $0.133^{\pm.009}$  \\
 \checkmark & {\checkmark} & & {\checkmark} & ${0.778^{\pm.002}}$ & $0.119^{\pm.009}$ \\
 \checkmark & {\checkmark} & {\checkmark} & & $0.769^{\pm.004}$ & $0.107^{\pm.009}$ \\
\rowcolor{aliceblue!60} \checkmark & \checkmark & {\checkmark} & {\checkmark}  & $\bm{0.788^{\pm.002}}$ & $\bm{0.057^{\pm.006}}$ \\
\bottomrule[1.pt]
\\
\end{tabular}
}
\caption{\textbf{Evaluation of the motion VAE models on the motion part on the HumanML3D test set.}}
\label{apxtab: vae}
\resizebox{1.\linewidth}{!}{
\begin{tabular}{lccc}
\toprule[1.pt]
\multirow{2}{*}{{Methods}}  & \multirow{2}{*}{{Token Size}} & {R-Precision} & \multirow{2}{*}{{FID $\downarrow$}} \\
 &  & Top-3 $\uparrow$ & \\ 
\midrule
{Real Motion} & {-} & {$0.797^{\pm.002}$} & {$0.002^{\pm.000}$} \\ \midrule
Motion Level & 2 & $0.791^{\pm.003}$ & $1.906^{\pm.003}$ \\
Action Level & 4 & $0.793^{\pm.003}$ & $0.068^{\pm.002}$ \\
\rowcolor{aliceblue!60} Specific Level & 8 & $\bm{0.800}^{\pm.004}$ & $\bm{0.019}^{\pm.003}$ \\
\bottomrule[1.pt]
\end{tabular}
}
}
\end{minipage}
\hfill
\begin{minipage}[t]{0.48\textwidth}
\centering
\caption{\textbf{Effect of diffusion steps on the HumanML3D test set.} We use DDIM in practice and set $T^m$, $T^a$, and $T^s$ to 50 for optimal performance.}
\label{tab:ab2}
\resizebox{1.\linewidth}{!}{
\begin{tabular}{lcccc}
\toprule[1.pt]
\multirow{2}{*}{{Methods}}  & \multicolumn{3}{c}{{Diffusion Steps}} & \multirow{2}{*}{{FID $\downarrow$}} \\
\cmidrule(rl){2-4}
& $T^m$ & $T^a$ & $T^s$  \\ 
\midrule
\multicolumn{5}{l}{\emph{\color{gray}{\textbf{1000 diffusion steps with DDPM~\cite{ho2020denoising}}}}} \\
MDM~\cite{tevet2022human}~\pub{ICLR23} & 1000 & {\color{gray}\ding{56}} & {\color{gray}\ding{56}} & $0.544^{\pm.044}$  \\
MotionDiffuse~\cite{zhang2022motiondiffuse}~\pub{TPAMI24} & 1000 & {\color{gray}\ding{56}} & {\color{gray}\ding{56}}  & $0.630^{\pm.001}$ \\
\midrule
\multicolumn{5}{l}{\emph{\color{gray}{\textbf{50 diffusion steps with DDIM~\cite{song2020denoising}}}}} \\
MLD~\cite{chen2022executing}~\pub{CVPR23} & 50 & {\color{gray}\ding{56}} & {\color{gray}\ding{56}} & $0.473^{\pm.013}$ \\ 
\rowcolor{aliceblue!60} GuidedMotion (Ours) & 20 & 15 & 15  & $0.136^{\pm.007}$ \\
\rowcolor{aliceblue!60} GuidedMotion (Ours) & 15 & 20 & 15  & ${0.120^{\pm.006}}$ \\
\rowcolor{aliceblue!60} GuidedMotion (Ours) & 15 & 15 & 20  & $\bm{0.117^{\pm.006}}$ \\ \midrule
\multicolumn{5}{l}{\emph{\color{gray}{\textbf{150 diffusion steps with DDIM~\cite{song2020denoising}}}}} \\
MLD~\cite{chen2022executing}~\pub{CVPR23} & 150 & {\color{gray}\ding{56}} & {\color{gray}\ding{56}}  & $0.457^{\pm.011}$   \\ 
\rowcolor{aliceblue!60} GuidedMotion (Ours) & 50 & 50 & 50  & $\bm{0.057^{\pm.006}}$ \\ \midrule
\multicolumn{5}{l}{\emph{\color{gray}{\textbf{300 diffusion steps with DDIM~\cite{song2020denoising}}}}} \\
MLD~\cite{chen2022executing}~\pub{CVPR23} & 300 & {\color{gray}\ding{56}} & {\color{gray}\ding{56}} & $0.403^{\pm.011}$   \\ 
\rowcolor{aliceblue!60} GuidedMotion (Ours) & 100 & 100 & 100 & $\bm{0.062^{\pm.007}}$ \\ 
\bottomrule[1.pt]
\end{tabular}
}
\end{minipage}
\hfill
\end{table}

\begin{figure}[h]
\centering
\includegraphics[width=1.\linewidth]{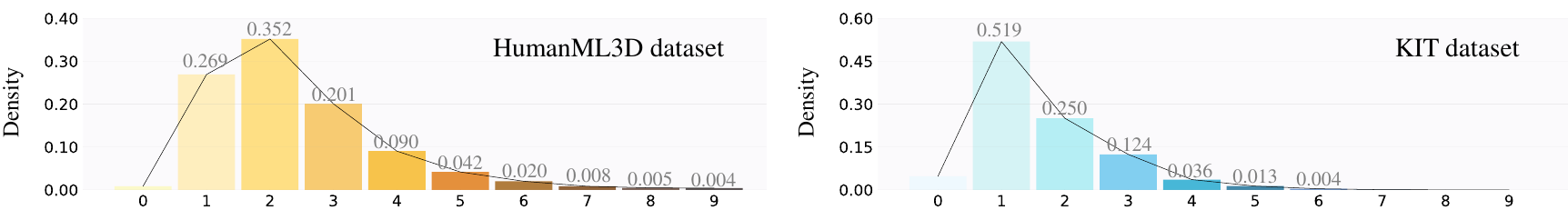}
\caption{\textbf{The distribution of the number of local actions in each motion.} Motions typically consist of multiple local actions rather than just one local action.}
\label{fig:distribution}
\end{figure}

\noindent \myparagraph{Evaluation on Complex Motion Generation.}
We analyze the distribution of the number of local actions in each motion in \cref{fig:distribution}. As shown in \cref{fig:distribution}, motions typically consist of multiple local actions. However, generating complex motions, characterized by many local actions, poses a challenge. Compared with the direct generation of complex motion, our local-to-global paradigm, utilizing local actions as a prior, simplifies the intricacies involved in generating complex motions. To demonstrate the advantages of the proposed local-to-global generation paradigm in generating complex motions, we create a new test set from the HumanML3D test set, consisting of motions with at least 3 local actions and lasting 150 frames or more. As shown in \cref{tab:complex}, our method maintains generation quality even for complex motions and is superior to other methods on most metrics. These results demonstrate the benefits of the proposed local-to-global paradigm in generating complex motions comprising multiple local actions.

\noindent \myparagraph{Ablative Analysis.}
\mysubparagraph{Analysis of each part of our method.} 
To explore the impact of each part of our method, we provide the ablation results in \cref{tab:ab1}. In the proposed hierarchical motion diffusion model, the high semantic layer generates results based on the results from the low semantic layer. As shown in \cref{tab:ab1}, higher semantic levels, such as the specific level, exhibit superior performance. Moreover, the proposed local action guidance significantly enhances the quality of the generated motion, providing conclusive evidence for the effectiveness of the proposed method. We observe that the performance in the ``R-Precision Top-3'' metric at the specific level, without local action guidance, is lower compared to the action level. This is likely due to the specific level refining results based on the action details. When two motion descriptions share overlapping action details, the model may produce similar features in the generated motions, thus adversely affecting R-Precision. Despite this, since the specific level can enhance the quality (FID) of motion generation, we still recommend its utilization.

\begin{figure*}[t]
\centering
\includegraphics[width=1\textwidth]{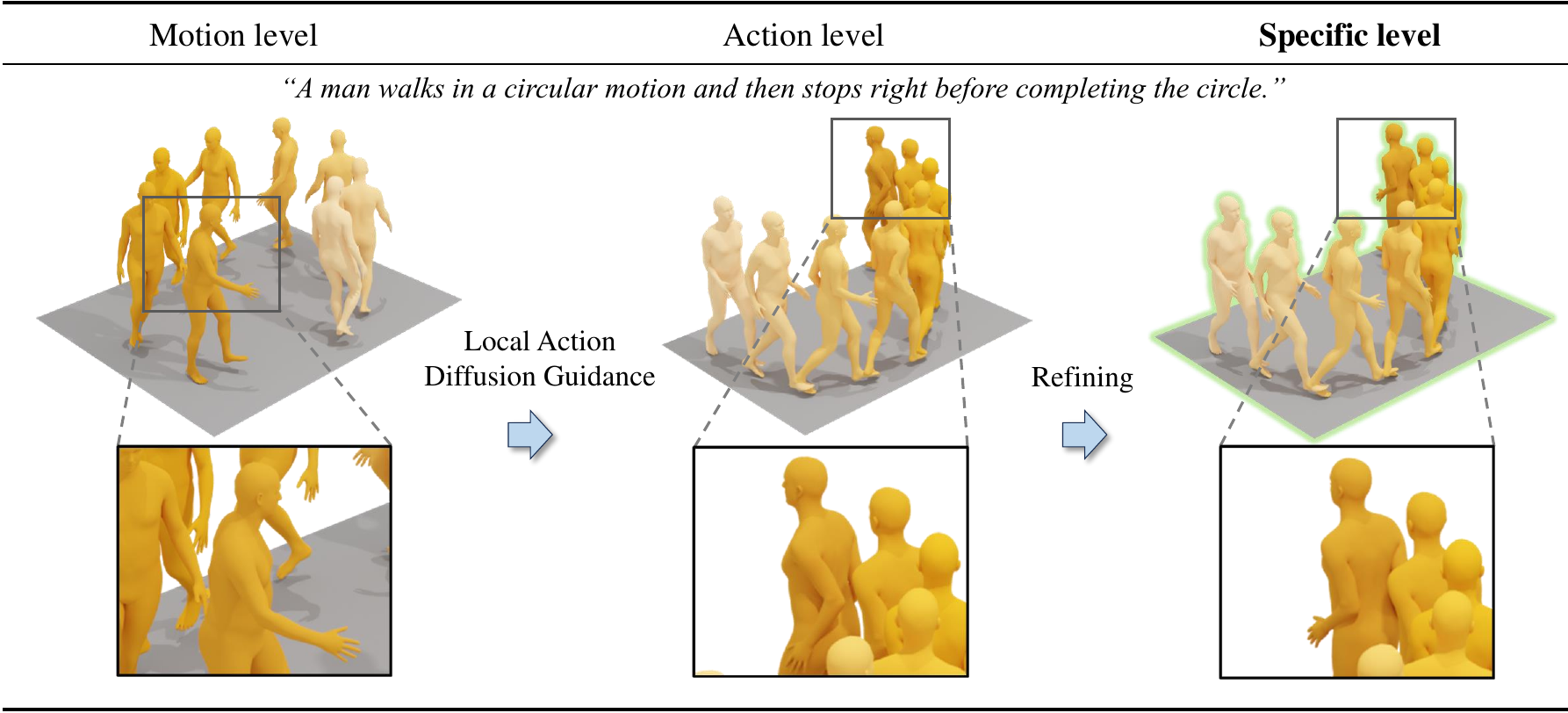}
\caption{\textbf{Qualitative comparison of different hierarchical levels.} The output at the higher level (\eg, specific level) contains more action details.}
\label{appendix:fig different hierarchies}
\end{figure*}

\noindent \mysubparagraph{Analysis of the motion VAE models.}
In \cref{apxtab: vae}, we show the evaluation of the motion VAE models on the HumanML3D test set. Among the three levels, the performance of the specific level is the best, which indicates that increasing the token size enhances the reconstruction ability of the motion VAE models. Therefore, we take the output at the specific level as the final result and use the motion decoder to decode the latent representation into the motion sequence. 

\noindent \mysubparagraph{Analysis of the diffusion steps.} 
In \cref{tab:ab2}, we provide the ablation results of the total number of diffusion steps on the HumanML3D test set. We observe that the number of diffusion steps at the higher level, such as the specific level, has a more pronounced impact on quantitative results. Simultaneously, the number of diffusion steps at the action level determines the control ability of the local action guidance to the global motion. Therefore, we recommend allocating a sufficient number of diffusion steps to each level. As illustrated in \cref{tab:ab2}, the performance is similar when the total number of diffusion steps is set to 150 and 300, prompting us to adopt a setting of 150 steps in practice.

\begin{figure*}[tbp]
\centering
\includegraphics[width=1\textwidth]{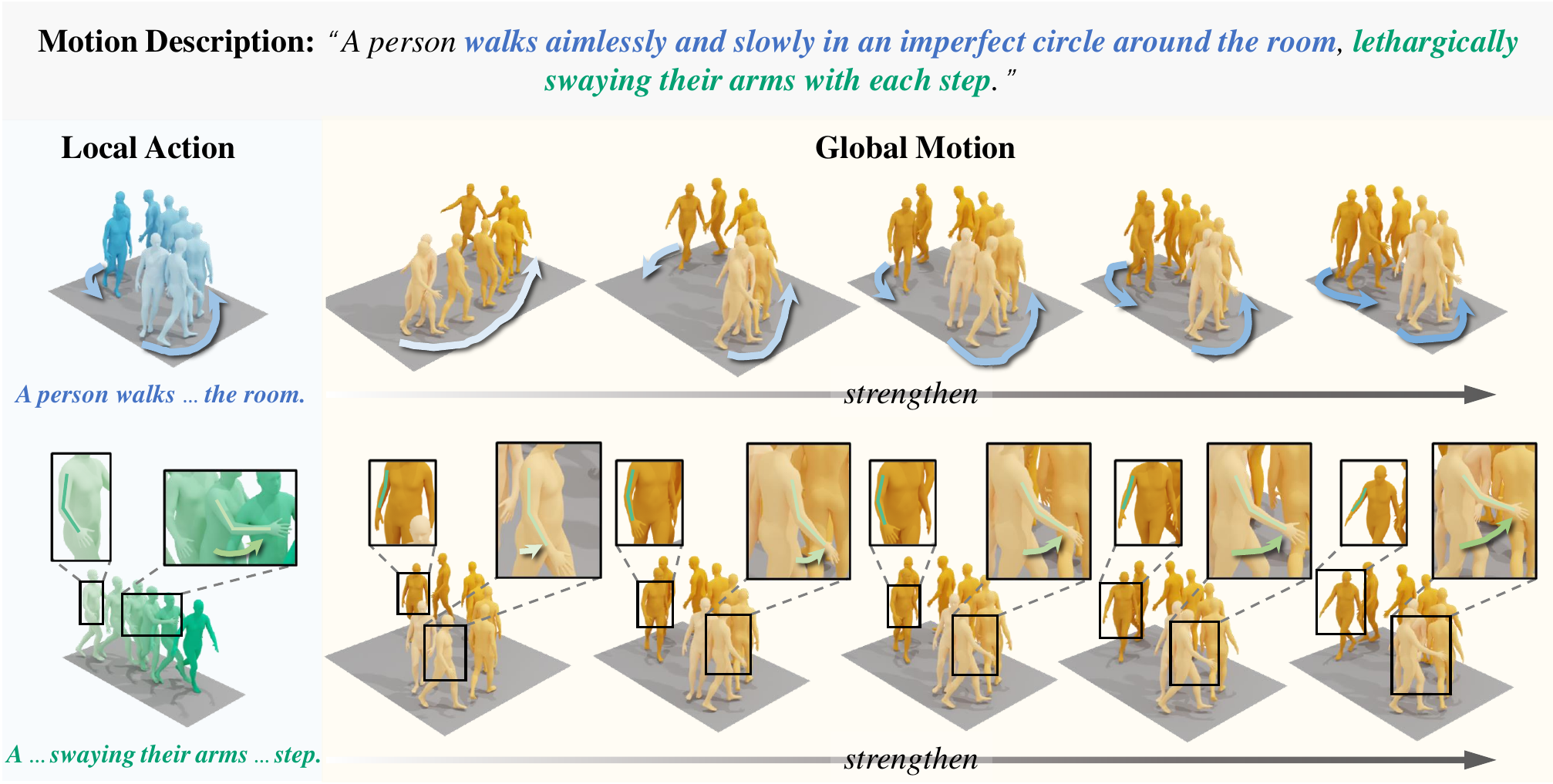}
\caption{\textbf{The proposed GuidedMotion controls the generation process of motion diffusion models.} Our method provides flexibility in adjusting the guiding weight $\lambda$ of each local action, enabling fine-grained control over global motion.}
\label{fig: fig0}
\end{figure*}

\noindent \myparagraph{Qualitative Analysis.} 
\noindent \mysubparagraph{Visualization of different hierarchies.} 
The results in \cref{appendix:fig different hierarchies} show that the output at the higher level (\eg, specific level) contains more action details. Specifically, the motion level generates only coarse-grained overall motion. The action level generates local actions with guidance but lacks action specifics. The specific level generates more action specifics than the action level.

\noindent \mysubparagraph{Visualization of adjusting the guiding weight of each local action.} 
The strength of our local action-guided motion generation lies in its capacity to fine-tune the generation process of motion diffusion models. In contrast to existing methods confined to producing a singular style of motion, our method offers flexibility in adjusting the guiding weight $\lambda$ of action guidance. This affords precise control over how each local action influences the overall motion, catering to diverse user preferences. As illustrated in \cref{fig: fig0}, we can manipulate the movement trajectories by varying the guiding weight of the local action. For example, increasing the guiding weight of ``\textit{walks aimlessly and slowly in an imperfect circle around the room}'' results in the human body walking in a tighter circle. Furthermore, we can refine the human body postures throughout the motion. For instance, by amplifying the guiding weight of ``\textit{lethargically swaying their arms with each step},'' the body exhibits more pronounced arm movements.

\section{Conclusion}
In this paper, we introduce GuidedMotion, a local action-guided motion diffusion model designed to enhance the controllability of text-driven human motion generation by employing local actions as fine-grained control signals. Our method empowers users to combine preferred local actions freely, generating motions that align with their mental imagery. Extensive experiments demonstrate that our method achieves superior controllability than the existing state-of-the-art methods. Furthermore, our method supports continuous guiding weight adjustment, allowing for the refinement of the motion results to align with user preferences.

\section*{Acknowledgements}
This work was supported in part by the National Key R\&D Program of China (No. 2022ZD0118101), Natural Science Foundation of China (No. 61972217, 32071459, 62176249, 62006133, 62271465, 62332002, 62202014), the Shenzhen Medical Research Funds in China (No. B2302037), and AI for Science (AI4S)-Preferred Program, Peking University Shenzhen Graduate School, China.

%
%
\bibliographystyle{splncs04}
\bibliography{main}

\clearpage
\renewcommand{\thefootnote}{\fnsymbol{footnote}}
\renewcommand{\thetable}{\Alph{table}}
\renewcommand{\theequation}{\Alph{equation}}
\renewcommand{\thefigure}{\Alph{figure}}

\setcounter{table}{0}
\setcounter{section}{0}
\setcounter{figure}{0}
\setcounter{equation}{0}

This appendix provides implementation details~(\cref{apx: Implementation Details}), several additional analyses~(\cref{apx: Additional Analyses}), additional discussions~(\cref{apx: Additional Discussions}), and details of motion representations and metric definitions~(\cref{apx: Motion Representations and Metric Definitions}).

\section{Implementation Details}\label{apx: Implementation Details}
\subsection{Semantic Graph Parsing}
To obtain actions, action attributes, and the semantic role of each attribute in relation to the corresponding action, we implement a semantic parser of motion descriptions based on a semantic role parsing toolkit~\cite{shi2019simple}\footnote{\href{https://allenai.org/allennlp}{https://allenai.org/allennlp}}. Specifically, the parser, when given a motion description, identifies the verbs within the sentence, extracts attribute phrases corresponding to each verb, and determines the semantic role of each attribute phrase. In the semantic graph, the entire sentence is treated as the global motion node. Verbs serve as action nodes and are connected to the motion node through direct edges, facilitating implicit learning of temporal relationships among different actions during graph reasoning. Attribute phrases represent specific nodes, linked to action nodes. The edge type between an action node and a specific node is determined by the semantic role of the specifics in relation to the action. As shown in Tab.~\ref{appendix:tab:graph}, we extract three types (motions, actions, and specifics) of nodes and twelve types of edges to depict various associations among these nodes.

\begin{table}[thb]
\centering
\setlength{\tabcolsep}{7.5pt}
\caption{\textbf{Node types and edge types in the semantic graph.} Each edge type corresponds to a type of semantic role.}
\label{appendix:tab:graph}
{
\begin{tabular}{ll}
\toprule[1.pt]
Node type & Description\\ \midrule
Motion & global motion description\\
Action & verb\\
Specific & attribute of action\\
\midrule[1.pt]
Edge type & Description\\ \midrule
ARG0 & agent\\
ARG1 & patient\\
ARG2 & instrument, benefactive\\
ARG3 & start point\\
ARG4 & end point\\
ARGM-LOC & location (where)\\
ARGM-MNR & manner (how)\\
ARGM-TMP & time (when)\\
ARGM-DIR & direction (where to/from)\\
ARGM-ADV & miscellaneous\\
ARGM-MA & motion-action dependencies\\
OTHERS & other argument types, \eg, action\\
\bottomrule[1.pt]
\end{tabular}
}
\end{table}

\begin{table}[t]
\centering
\caption{\textbf{Evaluation of Inference time costs on the HumanML3D test set.} We evaluate the average time per sample with different diffusion schedules and FID. ``$\downarrow$'' denotes that lower is better. ``{\color{gray}\ding{56}}'' denotes that this method does not apply this parameter. We set $T^m$, $T^a$, and $T^s$ to 50 for optimal performance. It is important to note that because local actions can be pre-generated, the inference time costs do not encompass the time taken for the pre-generation of local actions.}
\label{apxtab: time}
\setlength{\tabcolsep}{7.5pt}{
\begin{tabular}{lcccc}
\toprule[1.pt]
\multirow{2}{*}{{Methods}}  & \multicolumn{3}{c}{{Diffusion Steps}} & \multirow{2}{*}{{Time (s) $\downarrow$}} \\
\cmidrule(rl){2-4}
& $T^m$ & $T^a$ & $T^s$  \\ 
\midrule
\multicolumn{5}{l}{\emph{\color{gray}{\textbf{1000 diffusion steps with DDPM~\cite{ho2020denoising}}}}} \\
MDM~\cite{tevet2022human}~\pub{ICLR23} & 1000 & {\color{gray}\ding{56}} & {\color{gray}\ding{56}} & $178.7699$  \\
MotionDiffuse~\cite{zhang2022motiondiffuse}~\pub{TPAMI24} & 1000 & {\color{gray}\ding{56}} & {\color{gray}\ding{56}}  & $5.5045$ \\
\midrule
\multicolumn{5}{l}{\emph{\color{gray}{\textbf{50 diffusion steps with DDIM~\cite{song2020denoising}}}}} \\
MLD~\cite{chen2022executing}~\pub{CVPR23} & 50 & {\color{gray}\ding{56}} & {\color{gray}\ding{56}} & $0.7471$ \\ 
\rowcolor{aliceblue!60} GuidedMotion (Ours) & 20 & 15 & 15  & $0.8115$ \\
\rowcolor{aliceblue!60} GuidedMotion (Ours) & 15 & 20 & 15  & $0.7684$ \\
\rowcolor{aliceblue!60} GuidedMotion (Ours) & 15 & 15 & 20  & $0.7603$ \\ \midrule
\multicolumn{5}{l}{\emph{\color{gray}{\textbf{150 diffusion steps with DDIM~\cite{song2020denoising}}}}} \\
MLD~\cite{chen2022executing}~\pub{CVPR23} & 150 & {\color{gray}\ding{56}} & {\color{gray}\ding{56}}  & $2.4998$   \\ 
\rowcolor{aliceblue!60} GuidedMotion (Ours) & 50 & 50 & 50  & $2.5189$ \\ 
\bottomrule[1.pt]
\end{tabular}
}
\end{table}

\subsection{Implementation Details for Different Datasets}
Following MLD~\cite{chen2022executing}, we utilize a frozen text encoder of the CLIP-ViT-L-14~\cite{radford2021learning} model for text representation. The dimension of node representation is set to 768. The dimension of latent embedding is set to 256. 

For the motion variational autoencoder, motion encoder and decoder all consist of 9 layers and 4 heads with skip connection~\cite{ronneberger2015u}. To generate motion from coarse to fine step by step, we encode motion independently into three latent representation spaces $\bm{z}^m\in \mathbb{R}^{Q^m\times D'}$, $\bm{z}^a\in \mathbb{R}^{Q^a\times D'}$ and $\bm{z}^s\in \mathbb{R}^{Q^s\times D'}$, where the number of tokens gradually increases, \ie, $Q^m\leq Q^a \leq Q^s$. We set the token sizes $Q^m$ to 2, $Q^a$ to 4, and $Q^s$ to 8. 

Following previous works, our denoiser network is learned with classifier-free diffusion guidance~\cite{ho2022classifier}. The classifier-free diffusion guidance improves the quality of samples by reducing diversity in conditional diffusion models. Concretely, it learns both the conditioned and the unconditioned distribution (10\% dropout~\cite{srivastava2014dropout}) of the samples. Finally, we perform a linear combination in the following manner, which is formulated as:
\begin{equation}
\begin{aligned}
&{\widehat{\boldsymbol{\epsilon}^m} = \alpha \phi_{m}(\bm{z}^m, t^m, \mathcal{V}^{m}) + (1-\alpha)\phi_{m}(\bm{z}^m, t^m, \varnothing),}\\
&{\widehat{\boldsymbol{\epsilon}^a} = \alpha \phi_{a}(\bm{z}^a, t^a, [\mathcal{V}^{m},\mathcal{V}^{a},\bm{z}^{m}]) + (1-\alpha)\phi_{a}(\bm{z}^a, t^a, \varnothing),}\\ 
&{\widehat{\boldsymbol{\epsilon}^s} = \alpha \phi_{s}(\bm{z}^s, t^s, [\mathcal{V}^{m},\mathcal{V}^{a},\mathcal{V}^{s},\bm{z}^{a}]) + (1-\alpha)\phi_{s}(\bm{z}^s, t^s, \varnothing)},
\end{aligned}
\end{equation}
Where $\alpha$ is the guidance scale and $\alpha>1$ can strengthen the effect of guidance~\cite{chen2022executing}. We set $\alpha$ to 7.5 in practice following MLD.

All our models are trained with the AdamW~\cite{kingma2014adam,loshchilov2017fixing} optimizer using a fixed learning rate of 1e-4. We use 4 Tesla V100 GPUs for the training, and there are 128 samples on each GPU, so the total batch size is 512. The number of diffusion steps of each level is 1,000 during training, and the step sizes $\beta_{t}$ are scaled linearly from 8.5$\times$1e-4 to 0.012. We keep running a similar number of iterations on different data sets. For the HumanML3D dataset, the model is trained for 6,000 epochs during the motion variational autoencoder stage and 3,000 epochs during the diffusion stage. For the KIT dataset, the model is trained for 30,000 epochs during the motion variational autoencoder stage and 15,000 epochs during the diffusion stage. 

\begin{table}[t]
\centering
\caption{\textbf{Evaluation of the motion VAE models on the motion part.} ``$\uparrow$'' denotes that higher is better. ``$\downarrow$'' denotes that lower is better. ``$\rightarrow$'' denotes that results are better if the metric is closer to the real motion. Among the three levels, the performance of the specific level is the best.}
\label{apxtab: vae2}
\setlength{\tabcolsep}{3.5pt}
{
\begin{tabular}{lcccccc}
\toprule[1.pt]
\multirow{2}{*}{{Methods}}  & \multirow{2}{*}{{Token Size}} &\multicolumn{3}{c}{{R-Precision $\uparrow$}} & \multirow{2}{*}{{FID $\downarrow$}} & \multirow{2}{*}{{Diversity $\rightarrow$}} \\
\cmidrule(rl){3-5}
 &  & Top-1 & Top-2 & Top-3 \\ 
\midrule
\multicolumn{7}{l}{\emph{\color{gray}{\textbf{VAE models on the HumanML3D dataset}}}} \\
\color{gray}{Real Motion} & \color{gray}{-} & \color{gray}{$0.511$} & \color{gray}{$0.703$} & \color{gray}{$0.797$} & \color{gray}{$0.002$} & \color{gray}{$9.503$} \\
Motion Level & 2 & $0.498$ & $0.692$ & $0.791$ & $1.906$ & $9.675$ \\
Action Level & 4 & $0.514$ & $0.703$ & $0.793$ & $0.068$ & $\bm{9.610}$ \\
\rowcolor{aliceblue!60} Specific Level & 8 & $\bm{0.525}$ & $\bm{0.708}$ & $\bm{0.800}$ & $\bm{0.019}$ & $9.863$ \\
\midrule
\multicolumn{7}{l}{\emph{\color{gray}{\textbf{VAE models on the KIT dataset}}}} \\
\color{gray}{Real Motion} & \color{gray}{-} & \color{gray}{$0.424$} & \color{gray}{$0.649$} & \color{gray}{$0.779$} & \color{gray}{$0.031$} & \color{gray}{$11.08$} \\ 
Motion Level & 2 & $\bm{0.431}$ & $0.623$ & $0.745$ & $1.196$ & $10.66$ \\
Action Level & 4 & $0.413$ & $\bm{0.644}$ & $\bm{0.770}$ & $0.396$ & $10.85$ \\
\rowcolor{aliceblue!60} Specific Level & 8 & $0.414$ & $0.640$ & $0.760$ & $\bm{0.361}$ & $\bm{10.86}$ \\
\bottomrule[1.pt]
\end{tabular}
}
\end{table}

\section{Additional Analyses}\label{apx: Additional Analyses}
\subsection{Analysis of the Inference Time}
In \cref{apxtab: time}, we provide the evaluation of inference time costs on the HumanML3D test set. It is important to note that because local actions can be pre-generated, the inference time costs presented in \cref{apxtab: time} do not encompass the time taken for the pre-generation of local actions. As shown in \cref{apxtab: time}, despite decomposing the diffusion process into three parts, our method demonstrates efficiency comparable to one-stage diffusion methods during the inference stage. This is achievable by controlling the total number $T^m+T^a+T^s$ of iterations, ensuring it aligns with that of one-stage diffusion methods. Our method exhibits inference speeds comparable to existing methods with the same total number of diffusion steps, underscoring the efficiency of our method.

\subsection{Analysis of the motion VAE models}
We provide the evaluation of the motion VAE models. In \cref{apxtab: vae2}, we show the results on the HumanML3D test set and KIT test set. Among the three levels, the performance of the specific level is the best, which indicates that increasing the token size enhances the reconstruction ability of the motion VAE models. We take the output at the specific level as the final result and use the motion decoder to decode the latent representation into the motion sequence. 

\begin{figure*}[t]
\centering
\includegraphics[width=1\textwidth]{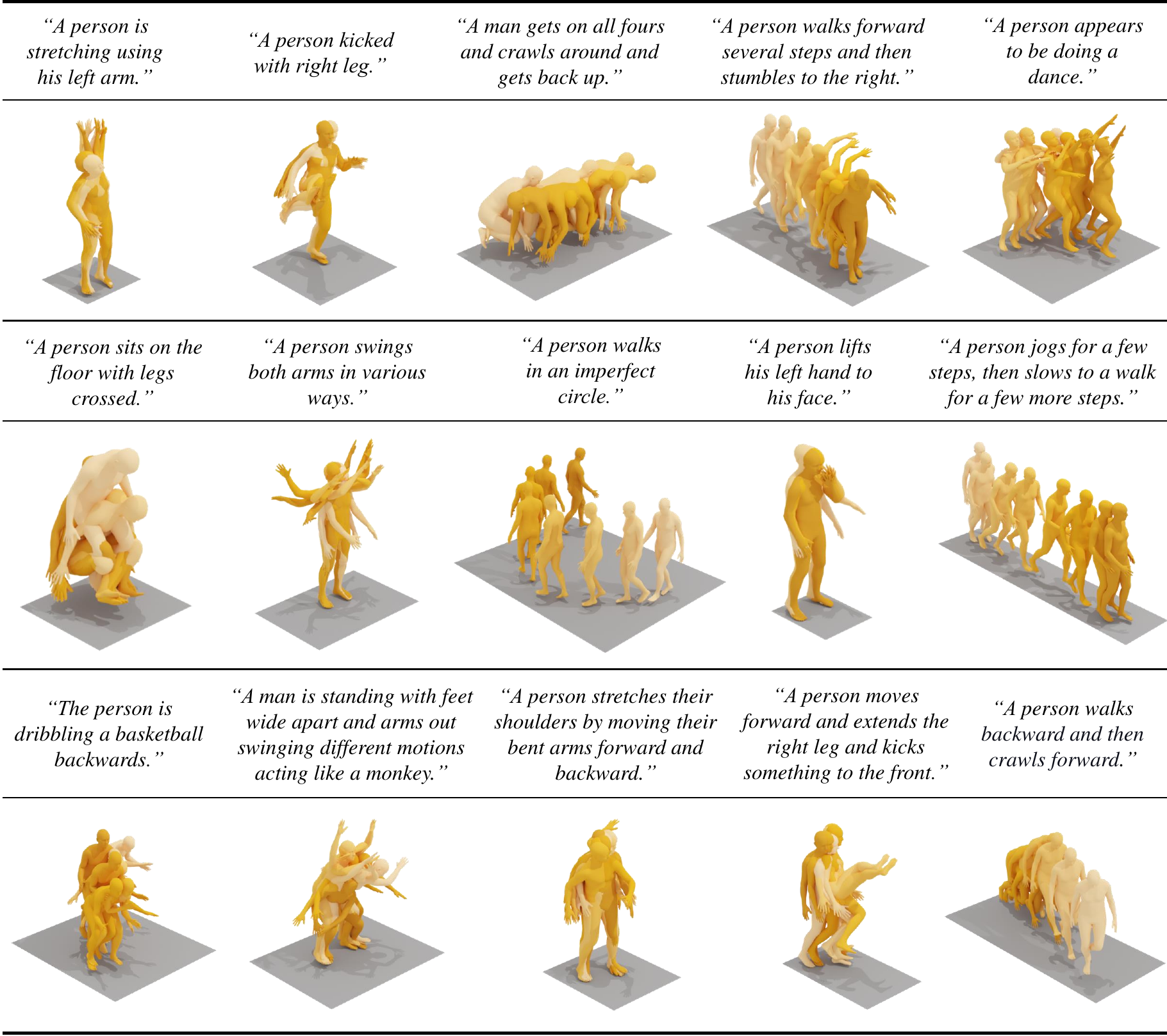}
\caption{\textbf{Additional qualitative motion results are generated with text prompts.} The darker colors indicate the later in time. These results demonstrate that our method can generate diverse and accurate motion sequences.}
\label{appendix:fig:more samples}
\end{figure*}

\subsection{Additional Visualization Results} 
In \cref{appendix:fig:more samples}, we provide additional qualitative motion results. These results demonstrate that our method can generate diverse and accurate motion sequences.

\section{Additional Discussions}\label{apx: Additional Discussions}
\subsection{Limitations of our Work} 
While our method has shown notable advancements, there exist several limitations that warrant further investigation. (1)~Although our method can generate results of arbitrary lengths, it still adheres to the maximum length observed in the dataset. Modeling a continuous human motion with temporal consistency introduces a fascinating challenge. (2)~Given that our method employs a diffusion process in the motion latent space rather than directly on the raw motion sequences, it is better suited for high-level motion editing rather than low-level motion editing, such as modifying the position of a single joint. Exploring low-level motion editing within latent space holds great promise and poses an exciting avenue for future research. (3)~Our method inherits the inherent randomness of diffusion models. While this trait enriches diversity, it is crucial to recognize that it may occasionally result in less desirable outcomes. (4)~The human motion synthesis capabilities of our method are constrained by the performance of the pre-trained motion variational autoencoders, which we discuss in experiments (\cref{apxtab: vae2}). This defect also exists in the existing methods, such as MLD~\cite{chen2022executing} and T2M-GPT~\cite{zhang2023t2m}, which also use motion variational autoencoder. Furthermore, delving into the realm of a more efficient motion latent space holds significant promise as a compelling avenue for future research. (5)~Though our method brings negligible extra cost to computations, it is still limited by the slow inference speed of existing diffusion models. However, with the development of diffusion models, we anticipate a progressive mitigation of this limitation. We discuss the inference time in \cref{apxtab: time}. This defect also exists in the existing state-of-the-art methods, such as MDM~\cite{tevet2022human}, MLD~\cite{chen2022executing}, GraphMotion~\cite{jin2023act}, and ReMoDiffuse~\cite{zhang2023remodiffuse}, which also use diffusion models.

\subsection{Future Work} 
In this work, we focus on improving the controllability of text-driven human motion generation. Our method empowers users to combine preferred local actions freely, thereby generating motions that resonate with their mental imagery. In future research, we aim to explore additional avenues for controlling motion generation. Particularly, delving into low-level motion editing within the latent space holds significant promise and presents an exciting direction.

\section{Motion Representations and Metric Definitions}\label{apx: Motion Representations and Metric Definitions}
\subsection{Motion Representations}
Motion representation can be summarized into the following four categories, and we follow the previous work of representing motion in latent space.

\noindent \myparagraph{Latent Format.}
Following previous works~\cite{petrovich2022temos,chen2022executing,zhang2023t2m}, we encode the motion into the latent space with a motion variational autoencoder~\cite{kingma2013auto}.

\noindent \myparagraph{HumanML3D Format.}
HumanML3D~\cite{guo2022generating} proposes a motion representation $\bm{x}^{1:L}$ inspired by motion features in character control. Specifically, the $i_{th}$ pose $\bm{x}^i$ is defined by a tuple consisting of the root angular velocity $r^a\in \mathbb{R}$ along the Y-axis, root linear velocities ($r^x,r^z\in \mathbb{R}$) on the XZ-plane, root height $r^y\in \mathbb{R}$, local joints positions $\bm{j^p}\in \mathbb{R}^{3N_j}$, velocities $\bm{j^v}\in \mathbb{R}^{3N_j}$, and rotations $\bm{j^r}\in \mathbb{R}^{6N_j}$ in root space, and binary foot-ground contact features $\bm{c^f}\in \mathbb{R}^4$ obtained by thresholding the heel and toe joint velocities. Here, $N_j$ denotes the joint number. Finally, the HumanML3D format can be defined as:
\begin{equation}
\bm{x}^i=\{r^a,r^x,r^z,r^y,\bm{j^p},\bm{j^v},\bm{j^r},\bm{c^f}\}.
\end{equation}

\noindent \myparagraph{SMPL-based Format.}
SMPL~\cite{loper2015smpl} is one of the most widely used parametric human models. SMPL and its variants propose motion parameters $\theta$ and shape parameters $\beta$. $\theta\in \mathbb{R}^{3\times23+3}$ is rotation vectors for 23 joints and a root, while $\beta$ represents the weights for linear blended shapes. The global translation $r$ is also incorporated to formulate the representation as follows:
\begin{equation}
\bm{x}^i=\{r,\theta,\beta\}.
\end{equation}

\noindent \myparagraph{MMM Format.}
Master Motor Map~\cite{terlemez2014master} (MMM) representations propose joint angle parameters based on a uniform skeleton structure with 50 degrees of freedom (DoFs). In text-to-motion tasks, recent methods~\cite{ahuja2019language2pose,ghosh2021synthesis,petrovich2022temos} converts joint rotation angles into $J=21$ joint XYZ coordinates. Given the global trajectory $t_{root}$ and $p_m\in \mathbb{R}^{3J}$, the preprocessed representation is formulated as:
\begin{equation}
\bm{x}^i=\{p_m,t_{root}\}.
\end{equation}

\subsection{Metric Definitions}
Following previous works, we use the following five metrics to measure the performance of the model. Note that global representations of motion and text descriptions are first extracted with the
pre-trained network in~\cite{guo2022generating}.

\noindent \myparagraph{R-Precision.} Under the feature space of the pre-trained network in~\cite{guo2022generating}, given one motion sequence and 32 text descriptions (1 ground-truth and 31 randomly selected mismatched descriptions), motion-retrieval precision calculates the text and motion Top 1/2/3 matching accuracy. 

\noindent \myparagraph{Frechet Inception Distance~(FID).} We measure the distribution distance between the generated and real motion using FID~\cite{heusel2017gans} on the extracted motion features~\cite{guo2022generating}. The FID is calculated as:
\begin{equation}
\begin{aligned} 
\textrm{FID}=\Vert \mu_{gt}-\mu_{pred}\Vert^2-\textrm{Tr}(\Sigma_{gt}+\Sigma_{pred}-2(\Sigma_{gt}\Sigma_{pred})^{\frac{1}{2}}),
\end{aligned} 
\end{equation}
where $\Sigma$ is the covariance matrix. Tr denotes the trace of a matrix. $\mu_{gt}$ and $\mu_{pred}$ are the mean of ground-truth motion features and generated motion features.

\noindent \myparagraph{Multimodal Distance~(MM-Dist).} Given $N$ randomly generated samples, we calculate the average Euclidean distances between each text feature $f_t$ and the generated motion feature $f_m$ from that text. The multimodal distance is calculated as:
\begin{equation}
\textrm{MM-Dist}=\frac{1}{N}\sum^{N}_{i=1}\Vert f_{t,i}-f_{m,i}\Vert,
\end{equation}
where $f_{t,i}$ and $f_{m,i}$ are the features of the $i_{th}$ text-motion pair.

\noindent \myparagraph{Diversity.} All generated motions are randomly sampled to two subsets (\ie, $\{\bm{x}_1,\bm{x}_2,...,\bm{x}_{X_d}\}$ and $\{\bm{x}_1^{'},\bm{x}_2^{'},...,\bm{x}_{X_d}^{'}\}$) of the same size $X_d$. Then, we extract motion features~\cite{guo2022generating} and compute the average Euclidean distances between the two subsets:
\begin{equation}
\textrm{Diversity}=\frac{1}{X_d}\sum^{\bm{X}_d}_{i=1}\Vert \bm{x}_i-\bm{x}_i^{'}\Vert.
\end{equation}

\noindent \myparagraph{Multimodality (MModality).} We randomly sample a set of text descriptions with size $J_m$ from all descriptions. For each text description, we generate $2\times X_m$ motion sequences, forming $X_m$ pairs of motions. We extract motion features and calculate the average Euclidean distance between each pair. We report the average of all text descriptions. We define features of the $j_{th}$ pair of the $i_{th}$ text description as ($\bm{x}_{j,i}$, $\bm{x}_{j,i}^{'}$). The multimodality is calculated as:
\begin{equation}
\textrm{MModality}=\frac{1}{J_m \times X_m}\sum^{J_m}_{j=1}\sum^{X_m}_{i=1}\Vert \bm{x}_{j,i}-\bm{x}_{j,i}^{'}\Vert.
\end{equation}

\end{document}